# A New Pipeline For Generating Instruction Dataset via RAG and Self Fine-tuning

Chih-Wei Sung
*Computer Scicence and Communication Engineering*
*Providence University*
Taichung, Taiwan
g1112209@pu.edu.tw

Yu-Kai Lee
*Computer Scicence and Information Engineering*
*Providence University*
Taichung, Taiwan
s1091906@pu.edu.tw

Yin-Te Tsai
*Computer Scicence and Communication Engineering*
*Providence University*
Taichung, Taiwan
yttsai@pu.edu.tw

*Abstract*—With the rapid development of large language models (LLMs) in recent years, there has been an increasing demand for domain-specific Agents that can cater to the unique needs of enterprises and organizations. Unlike general models, which strive for broad coverage, these specialized Agents rely on focused datasets tailored to their intended applications. This research proposes a pipeline that leverages the power of LLMs and the Retrieval-Augmented Generation (RAG) related framework to construct high-quality instruction datasets for fine-tuning on specific domains using custom document collections. By ingesting domain-specific documents, the pipeline generates relevant and contextually appropriate instructions, thus effectively creating a comprehensive dataset for fine-tuning LLMs on the target domain. This approach overcomes the limitations of traditional dataset creation methods, which often rely on manual curation or web-scraping techniques that may introduce noise and irrelevant data. Notably, our pipeline offers a dynamic solution that can quickly adapt to updates or modifications in the domain-specific document collection, eliminating the need for complete retraining. Additionally, it addresses the challenge of data scarcity by enabling the generation of instruction datasets from a limited set of initial documents, rendering it suitable for unpopular or specialized domains where comprehensive datasets are scarce. As a case study, we apply this approach to the domain of psychiatry, a field requiring specialized knowledge and sensitive handling of patient information. The resulting fine-tuned LLM demonstrates showcases the viability of the proposed approach and underscores its potential for widespread adoption across various industries and domains where tailored, accurate, and contextually relevant language models are indispensable.

## I. INTRODUCTION

Large language models (LLMs) have revolutionized natural language processing, yet face challenges when applied to highly specialized domains. For example, psychiatry requires nuanced understanding of technical terminology, diagnostic criteria, and sensitive patient interactions. However, collecting and curating datasets specific to the domain can be arduous and time-consuming. This research proposes a pipeline that combines LLMs with the Retrieval-Augmented Generation (RAG)[1] related framework to construct domain-specific instruction datasets. By ingesting curated document collections (e.g., academic papers, clinical guidelines), the pipeline harnesses the generative capabilities of LLMs and the information retrieval strengths of RAG. This iterative process yields comprehensive datasets capturing domain intricacies while maintaining factual accuracy.

As a case study, we apply this approach to psychiatry, utilizing the "Desk Reference to the Diagnostic Criteria from DSM-5" as our primary source document. The DSM-5 (Diagnostic and Statistical Manual of Mental Disorders, Fifth Edition)[2] is a crucial guideline book for psychiatric diagnosis, and the Desk Reference serves as a concise version of this comprehensive manual. We employ the Mistral-7B[3] model as our initial language model and leverage the Langchain framework[4] to implement the RAG technique, effectively imbuing the model with knowledge from the DSM-5.

Subsequently, through our carefully designed prompts, we facilitate the language model to generate relevant question-answer pairs in the desired format and content, which we then integrate into an instruction dataset. Finally, we fine-tune our Mistral-7B model using this dataset, transforming it into a specialized agent for the psychiatric domain, dubbed Mistral-7B-dsm5.

In the following sections, we will delve into the intricacies of designing this instruction dataset and discuss our approach to evaluating the performance of Mistral-7B-dsm5 in the psychiatric domain compared to ChatGPT. By employing GPT-4 as an expert judge[5], we aim to validate the effectiveness and viability of our proposed pipeline, demonstrating its potential for creating tailored LLMs across various specialized industries.

## II. RELATED WORK

### A. Self-generate instruction dataset

The success of ChatGPT has popularized the use of Instruction Tuning [6], subsequently validated by related research as the most effective fine-tuning approach for question-answering tasks [7]. However, the lack of available instruction datasets across various domains has posed a significant challenge. To address this issue, the Alpaca model from Stanford University introduced a self-instruction methodology [8, 9]. This approach involves initially curating a small seed pool of human-written instructions, which are then used as prompts to generate a larger instruction dataset via a LLM. This innovative solution has provided the academic community with a novel approach to overcome the limitations imposed by the scarcity of instruction datasets in specific domains.

*B. Downstream Tasks Application Framework*

In the realm of downstream task applications, two notable frameworks have emerged: Langchain and llama-index. These frameworks aim to facilitate the integration of language models into various domain-specific tasks and workflows.

Langchain is a comprehensive framework provides a modular and extensible approach to building applications with large language models. It offers a wide range of tools and utilities, including agents, memory components, and prompting techniques. Langchain allows developers to create custom pipelines and workflows, enabling seamless interaction between language models and external data sources, such as databases, APIs, and document repositories.

On the other hand, llama-index[10] is a specialized framework focused on building structured indices over unstructured data. It leverages language models to extract and organize information from text documents, enabling efficient retrieval and question-answering capabilities. llama-index supports various indexing strategies, including tree-based and vector-based approaches, allowing for flexible and optimized index structures tailored to specific use cases.

While both frameworks share the common goal of integrating language models into downstream tasks, they differ in their primary focus and approach. Langchain offers a broader set of tools for building end-to-end applications, while llama-index specializes in indexing and information retrieval from unstructured data sources.

In this research, we have adopted the Langchain framework due to its comprehensive and modular nature. Langchain's flexibility and extensive toolset align well with our proposed pipeline, enabling seamless integration of the Retrieval-Augmented Generation (RAG) approach and the generation of domain-specific instruction datasets. By leveraging Langchain, we can effectively combine the capabilities of large language models with the retrieval and augmentation processes required for our methodology.

*C. Retrieval-Augmented Generation*

The Retrieval-Augmented Generation (RAG) technique represents a significant advancement in enhancing the capabilities of LLMs by incorporating external knowledge sources. Originally introduced to enable LLMs to perform web searches and incorporate relevant information from the internet, RAG has since evolved to leverage custom document collections tailored to specific domains and downstream tasks. The core principle of RAG involves two key components: a retriever and a generator. The retriever component is responsible for identifying and retrieving relevant information from a given corpus of documents or knowledge base. This can be achieved through various techniques, such as sparse retrieval using TF-IDF or dense retrieval leveraging neural encoders to embed documents and queries into a shared vector space. The retrieved information is then passed to the generator component, typically an LLM, which generates contextually relevant and coherent responses by integrating the retrieved knowledge. This augmentation process allows the LLM to produce output that is not only fluent and grammatically correct but also factually grounded in the provided document collection.

The application of RAG in downstream task frameworks has further extended its capabilities. By integrating RAG with frameworks like Langchain, researchers and developers can seamlessly incorporate custom document collections into the retrieval process. This enables the generation of highly specialized and context-aware responses, tailored to the unique requirements of various domains and applications.

In our research, we leverage the RAG approach in conjunction with custom document collections to generate domain-specific instruction datasets. By retrieving relevant information from curated sources and incorporating it into the generation process, we can create instruction samples that accurately capture the nuances and intricacies of the target domain, while ensuring factual accuracy and contextual relevance.

*D. Large Language Model on Mental Healthcare*

Recent studies have explored the integration of LLMs with the field of mental healthcare to assist in maintaining patients' psychological well-being. Singh et al. [11] implemented a simple chatbot on mental health topics using the Langchain framework. Additionally, the Mental-LLM proposed by Xu et al. [12] utilizes online text data for the prevention of mental disorders. Their research also evaluated the predictive performance of various LLM models in this context.

Another noteworthy research is the Chat Doctor [13], a model based on LLaMA and trained on 100,000 dialogues between psychiatric patients and doctors. Although it does not use an instruction dataset but rather real-world dialogue, the Chat Doctor has shown promising accuracy in diagnosing mental health conditions. While not following the instruction-tuning approach, this model highlights the potential of LLMs in the mental healthcare domain when trained on substantial domain-specific data.

III. METHODOLOGY

Inspired by the work from Microsoft [14], we propose a pipeline that enables the generation of instruction datasets and self-fine-tuning for specialized domains. In contrast to traditional dataset construction methods, our approach leverages the power of RAG combined with LLMs, offering two advantages:

- Dynamic: If updates or modifications are required for the domain-specific documents, the RAG framework can quickly assist the model in ingesting new information. This dynamic nature represents a significant improvement over traditional methods, as it eliminates the need for complete retraining and fine-tuning, thereby reducing computational costs.

- Addressing Data Scarcity: In professionalized or specialized domains, obtaining comprehensive datasets can be challenging. Our pipeline alleviates this issue by enabling the generation of instruction datasets from a limited set of initial documents, making it suitable for domains with scarce data.

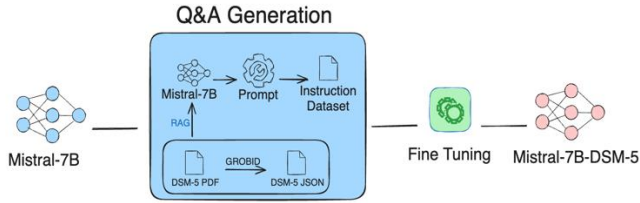

Fig. 1. Methodology pipeline. First, the DSM-5 PDF is converted into DSM-5 JSON using GROBID. Then, the instruction dataset obtains information from the DSM-5 JSON through RAG and generates a QAdataset using prompts. The generated question-answer pairs are subsequently utilized for fine-tuning the Mistral-7B.

The flowchart in Figure 1 illustrates the process of our proposed methodology. The following sections will delineate it step-by-step:

*A. Data Collecting and Preprocessing*

In the field of psychiatry, the DSM-5 (Diagnostic and Statistical Manual of Mental Disorders, Fifth Edition) is a widely recognized standard guideline. For our research, we utilized the more concise version, "Desk Reference to the Diagnostic Criteria from DSM-5," a 444-page PDF document.

However, instead of directly loading the PDF using a unstructured loader, we aimed to enhance the model's question-answering accuracy by transforming the unstructured text into a structured JSON format. To achieve this, we employed GROBID[15], a machine learning model designed for text extraction and structuring from PDF documents. GROBID extracts the raw text in TEI format, and then we convert it from XML to JSON file. By preprocessing the PDF using GROBID, we obtained a structured representation of the diagnostic criteria, disorder descriptions, and other relevant information from the DSM-5 desk reference. This structured data format facilitated more efficient information retrieval and improved the model's ability to provide accurate and contextually relevant responses within the psychiatric domain.

*B. Language Model and RAG setup*

We selected the Mistral 7B-instruct v0.2 language model for our research, as it outperforms other models of similar parameter size. We then leveraged the Langchain framework to ingest and process the structured JSON file obtained from the preprocessing step.

Within the Langchain framework, we employed the Conversational Retrieval Chain function, which is tailored for conversational and contextual retrieval tasks. For the vector database, we opted for FAISS (Facebook AI Similarity Search) [16], a highly efficient and scalable library for similarity search and dense vector indexing.

Regarding the choice of embedding model, we extensively evaluated various options, Ultimately, we selected the Nomic AI's embedding model which called "nomic-embed-text" [17], as it demonstrated superior performance compared to the other alternatives in our experiments.

*C. Generate QA Prompt Design*

Prompt design is a crucial part of our methodology, as it directly impacts the language model's ability to comprehend and generate relevant dataset entries accurately. To ensure high-quality datasets that encompass all essential information from the document corpus, we adopted a structured approach based on the table of contents.

We divided the prompts into sections according to the organization of the source material and generated 60~100 entries for each category of mental disorders, After deduplication and removal of low-quality entries, we targeted approximately 80 entries per category of mental disorders for dataset construction. This systematic approach aimed to capture the nuances and specific details associated with different psychiatric conditions. An example of our prompt design was shown in Figure 2.

> "Extract professional knowledge about {Disorder category} from the medical document and organize it in question-and-answer format. Each question should address a specific aspect of the topic to ensure comprehensiveness and clarity. Please output in the following JSON format :\
>
> {\
> "instruction": {QUESTION}, \
> "output": {ANSWER} \
> }"

Fig. 2. QA generation prompt example. When prompting the model, we replace the {Disorder category} tag with the title of each chapter in the DSM-5.

*D. Training*

After the aforementioned processes, we generated approximately 2,000 entries for our instruction fine-tuning dataset. The details of this dataset will be elaborated in Chapter 4. Subsequently, we utilized this dataset to fine-tune the Mistral-7B model.

For the fine-tuning process, we employed the LoRA (Low-Rank Adaptation) technique [18], which is a popular and efficient method for adapting large language models. The hardware used for training was a single RTX A6000 GPU with 48GB of VRAM. The total training time was 2 hours. The remaining hyperparameters and settings are summarized in Table I. Through this training process, we obtained the Mistral-7B-DSM-5 model, a language model tailored for the psychiatric domain, capable of performing simple diagnoses and psychological assessments.

TABLE I. HYPERPARAMETER CONFIGURATION

| Hyper parameter | Value |
|---|---|
| Learning Rate | 5e-5 |
| Lora r | 16 |
| Batch size | 2 |
| Epochs | 10 |
| Gradient accumulation | 8 |
| LR scheduler | cosine |

## IV. EXPERIMENT RESULTS

After completing the pipeline described in the previous chapter, we obtained a DSM-5 instruction dataset. This dataset comprises approximately 80 entries for each category of mental disorders, covering a range of relevant questions and answers. However, for the sections on "Medication-induced Movement" and "Other Adverse Effects of Medication", due to their relatively limited content, we generated 40 entries for each. To further enhance the diversity of the dataset, we additionally generated over 300 entries based on the entire DSM-5 guideline. The final dataset consists of approximately 2,000 instruction entries in total. The distribution of entries across different categories of mental disorders is presented in Table II.

This comprehensive instruction dataset serves as the foundation for fine-tuning our language model, aiming to enhance its performance in addressing psychiatric domain-specific tasks and queries accurately. The diversity of the dataset, encompassing various mental disorders and their associated diagnostic criteria, treatment recommendations, and other relevant information, is crucial for developing a versatile model capable of handling a wide range of scenarios within the psychiatric field.

TABLE II. DSM-5 INSTRUCTION DATASET CATEGORY PERCENTAGE

| Disorders Category | Percent(%) |
|---|---|
| Neurodevelopmental | 4% |
| Schizophrenia Spectrum and other Psychotic | 4% |
| Bipolar and Related | 4% |
| Depressive | 4% |
| Anxiety | 4% |
| Obsessive-Compulsive and Related | 4% |
| Trauma-and Stressor-Related | 4% |
| Dissociative | 4% |
| Somatic Symptom and Related | 4% |
| Feeding and Eating | 4% |
| Elimination | 4% |
| Sleep-Wake | 4% |
| Sexual Dysfunctions | 4% |
| Gender Dysphoria | 4% |
| Disruptive, Impulse-Control, and Conduct | 4% |
| Substance-Related and Addictive | 4% |
| Neurocognitive | 4% |
| Personality | 4% |
| Paraphilic | 4% |
| Other Mental Disorders | 4% |
| Medication-induced Movement | 2% |
| Other Adverse Effects of Medication | 2% |
| Misc. | 16% |

On the other hand, our model achieved an average final Loss of around 0.07 after training, as shown in Figure 3. While it is possible that increasing the number of epochs or adjusting the learning rate could further reduce the Loss, the current model performance is already outstanding.

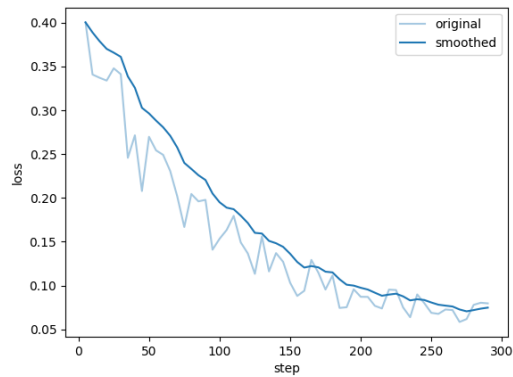

Fig. 3. Training loss. The stable decrease in training loss indicates that our model has effectively learned from the dataset.

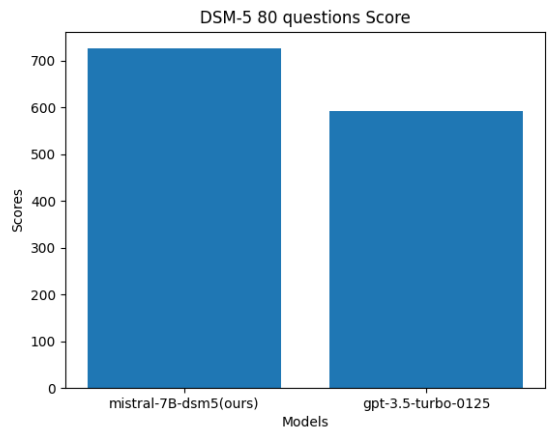

Fig. 4. Model evaluation scores judged by GPT-4.

At last, we employed GPT-4 as an expert judge to evaluate the capabilities of our fine-tuned model. We randomly generated 80 questions based on the "Desk Reference to the Diagnostic Criteria From DSM-5" guideline book, and tasked both our mistral-7B-dsm5 model and the GPT-3.5-turbo-0125 model to generate answers. These responses were then submitted to GPT-4 for scoring on a scale of 1 to 10 for each question. The results was shown in Figure 4, our mistral-7B-dsm5 model achieved a superior score of 726, while the GPT-3.5-turbo-0125 model scored 592.

## V. Conclusion

In this research, we have presented a novel pipeline that leverages the power of LLMs and the Retrieval-Augmented Generation related framework to generate domain-specific instruction datasets for fine-tuning. By conducting a case study in the field of psychiatry, we have demonstrated the feasibility and effectiveness of our approach. Remarkably, using only the electronic version of the DSM-5 guideline as the source document, our pipeline generated an instruction fine-tuning dataset comprising approximately 2,000 entries and 200,000 words. This achievement underscores the potential of our methodology to create high-quality, domain-specific datasets efficiently, even when working with limited source materials.

The success of our pipeline in the psychiatric domain serves as a compelling proof-of-concept, highlighting its applicability across various domains. This research provides enterprises and organizations with a viable solution for developing customized, domain-specific language models without incurring excessive costs. Notably, our experiments were conducted on consumer-grade hardware, further emphasizing the accessibility and scalability of our approach.

While our results are promising, we acknowledge that there is room for further improvement and refinement. In the future, we plan to extend our pipeline to different domains, continuously enhancing its usability and streamlining the process. By doing so, we aim to empower a broader range of stakeholders to leverage the power of LLMs for their specific needs, fostering innovation and unlocking new possibilities in natural language processing applications.

Moreover, our research underscores the importance of responsible and ethical practices in the development and deployment of AI systems, particularly in sensitive domains like mental healthcare. As language models become increasingly sophisticated, it is crucial to ensure their outputs are accurate, unbiased, and aligned with professional guidelines and ethical principles.

In conclusion, this study represents a significant step towards democratizing the creation of customized language models tailored to specific domains. By combining the strengths of LLMs, the RAG framework, and domain-specific knowledge sources, our pipeline offers a promising solution for organizations seeking to harness the potential of AI while addressing the challenges of data scarcity and domain specificity.